\documentclass[conference,letterpaper]{IEEEtran}

\usepackage[utf8]{inputenc} 
\usepackage[T1]{fontenc}
\usepackage{url}
\usepackage{ifthen}
\usepackage[cmex10]{amsmath} 

\usepackage[table]{xcolor}
\usepackage{dcolumn}
\newcolumntype{z}[1]{D{.}{.}{#1}}%

\usepackage{graphicx}
\usepackage{amsmath}
\usepackage{amssymb}
\usepackage{booktabs}
\usepackage{mathrsfs}
\usepackage{bbm}
\usepackage{multirow}

\usepackage{subfig}

\usepackage{amsfonts}      
\usepackage{nicefrac}      
\usepackage{microtype}     
\usepackage{algorithm}
\usepackage{algorithmic}
\usepackage{xspace}
\usepackage{bm}
\usepackage{enumitem}
\usepackage{hyperref}

\newcommand{\latinphrase}[1]{\emph{#1}}    %
\newcommand{\etal}{\latinphrase{et~al.}\xspace}
\newcommand{\ie}{\latinphrase{i.e.}\xspace}
\newcommand{\eg}{\latinphrase{e.g.}\xspace}

\newcommand{\yy}[0]{\ensuremath{\mathbf{y}}}

\newcommand{\bPhi}{\text{\boldmath $\Phi$}}

\usepackage{theorem}
\newcommand{\BlackBox}{\rule{1.5ex}{1.5ex}}  

\newtheorem{theorem}{Theorem}

\begin{document}
\title{Neural Group Testing to Accelerate Deep Learning}

\author{%

  \IEEEauthorblockN{Weixin Liang and James Zou}
  \IEEEauthorblockA{Stanford University, Stanford, CA 94305\\
                    Email: \{wxliang, jamesz\}@stanford.edu }
}


\maketitle

\begin{abstract}
   Recent advances in deep learning have made the use of large, deep neural networks with tens of millions of parameters. The sheer size of these networks imposes a challenging computational burden during inference. Existing work focuses primarily on accelerating each forward pass of a neural network. Inspired by the group testing strategy for efficient disease testing, we propose neural group testing, which accelerates by testing a group of samples in one forward pass. Groups of samples that test negative are ruled out. If a group tests positive,  samples in that group are then retested adaptively. A key challenge of neural group testing is to modify a deep neural network so that it could test multiple samples in one forward pass. We propose three designs to achieve this without introducing any new parameters and evaluate their performances. We applied neural group testing in an image moderation task to detect rare but inappropriate images. We found that neural group testing can group up to $16$ images in one forward pass and reduce the overall computation cost by over $73\%$ while improving detection performance. Our code is available at \url{https://github.com/Weixin-Liang/NeuralGroupTesting/}
\end{abstract}

\section{Introduction}

Recent advances in deep learning have been achieved by increasingly large and computationally-intensive deep neural networks. For example, a ResNeXt-101 network needs $16.51$ billion multiply-accumulate operations to process a single image~\cite{DBLP:journals/corr/CanzianiPC16}. Due to the challenging computation burden, deploying deep neural networks is expensive, and the cost scales linearly with the usage of the application. In addition, the inference cost is prohibitively expensive for privacy-sensitive applications that apply deep learning to encrypted data. The current state-of-the-art uses homomorphic encryption which makes each linear algebra step of the deep neural network very expensive to compute~\cite{encrptedData}. Inferring one image costs hundreds of seconds of GPU time. Recent studies also raise concern about the excessive energy consumption and $CO_2$ emission caused by running deep neural networks~\cite{DBLP:conf/acl/StrubellGM19,DBLP:journals/corr/abs-1907-10597}.

In this paper, we focus on the scenario where the class distribution is imbalanced. This scenario contains lots of real-world applications including image moderation (detecting rare but  inappropriate images)~\cite{moderator}, malware detection~\cite{malware,DBLP:conf/acsac/LiangBLLT18} and suicide prevention~\cite{suicide}. A key characteristic of these applications is that there is only a small portion of positives, and it is crucial to detect these positives.

Existing work on accelerating deep learning inference focuses primarily on accelerating each forward pass of the neural network. Specialized hardware designs reduce the time for each computation and memory access step of deep neural networks~\cite{hardware}. Parameter pruning reduces redundant parameters of the neural network that are not sensitive to the performance~\cite{han2015deep}. Neural architecture search automates the design of deep neural networks to trade-off between accuracy and model complexity~\cite{NAS}. To sum up, most existing methods reduce the time for running each forward pass of the neural network. In this paper, we explore an orthogonal direction that accelerates the neural network by reducing the number of forward passes needed to test a given amount of data.

\begin{figure}[tb]
\centering
\includegraphics[width=0.473\textwidth]{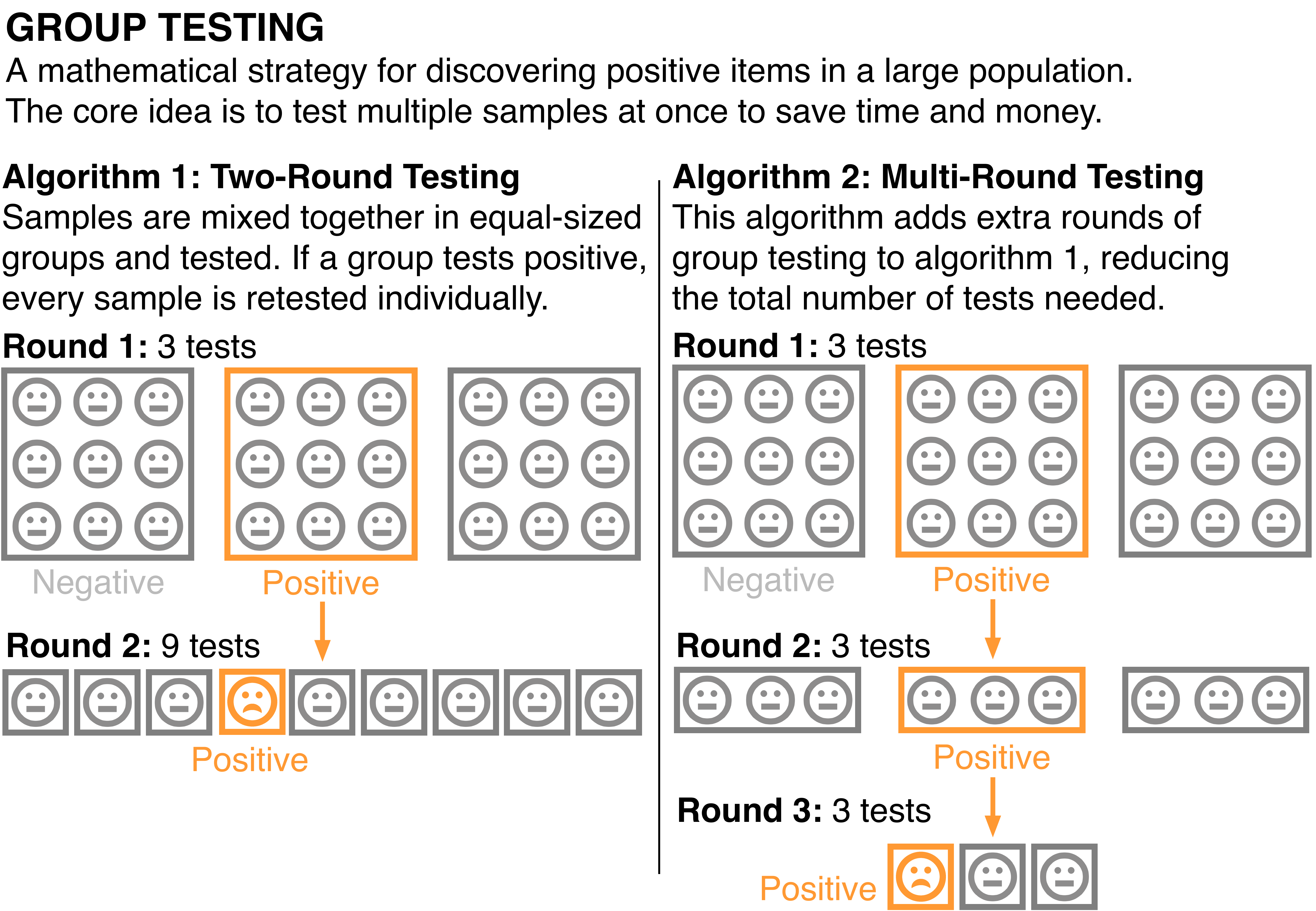} 
    \caption{
    Overview of group testing, a strategy that is used in efficient disease testing~\cite{mallapaty2020mathematical}. The core idea is to test multiple samples at once to save time and money. 
    } 
    \label{fig:algorithmSmall}
\end{figure}

\begin{figure}[tb]
\centering
\includegraphics[width=0.473\textwidth]{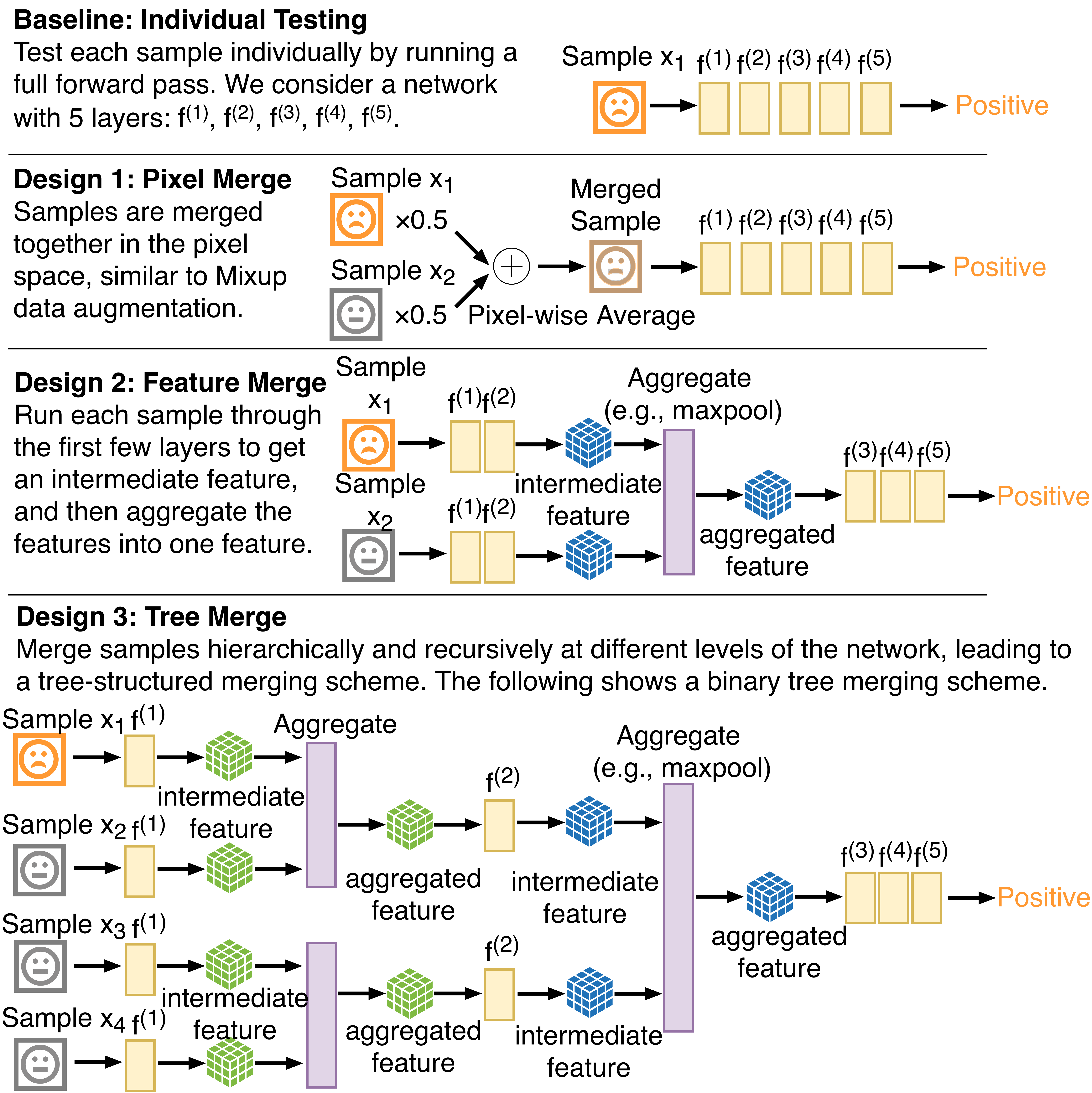} 
    \caption{
    Three neural network designs of neural group testing to test multiple samples with one forward pass. 
    } 
    \label{fig:Design}
\end{figure}

We propose neural group testing, a general deep learning acceleration framework that saves computation by testing multiple samples in one forward pass. Neural group testing combines deep learning and group testing, a strategy that is widely used for efficient disease testing~\cite{mallapaty2020mathematical}. As shown in Figure~\ref{fig:algorithmSmall}, the core idea of group testing is to pool samples from many people and test on the mixed sample. Groups of samples that test negative are ruled out, which saves testing many people individually. If a group tests positive, samples in that group are then retested adaptively. Similar to group testing, our framework reduces the number of tests (forward passes) of a neural network by allowing the network to test multiple samples in one forward pass, instead of testing each sample individually.

A key challenge of neural group testing is to modify a deep neural network so that it could test multiple samples in one forward pass. We achieve this without introducing any new parameters to the neural network. Our inspiration comes from \textit{Mixup}~\cite{mixup}, a data augmentation method that averages pairs of images in the pixel space as augmented training data. We propose three designs to address the challenge (See Figure~\ref{fig:Design}), starting from the most simple solution that merges samples by averaging their pixels and then tests on the merged sample. The proposed approach requires only several epochs of model fine-tuning. We applied neural group testing in an image moderation task to detect rare but inappropriate images. We found that neural group testing can group up to $16$ images in one forward pass and reduce the overall computation cost by over $73\%$ while improving detection performance. Because neural group testing can be performed without modifying the neural network architecture, it is complementary to and can be combined with all the popular approaches to speed up inference, such as sparsifying or quantizing networks or downsampling images. For example, we found that existing neural network pruning methods~\cite{pruning} can remove $80\%$ of the parameters in the neural network without any performance drop in neural group testing.

While the main contribution of our work is in the novel algorithm and empirical analysis,  it also motivates new questions and theories of group testing. Our work opens a window for applying group testing to unstructured data types, like images, that have been under-explored for group testing previously. The proposed method can be potentially applied to other unstructured media especially in deep learning settings. 

\section{Method} 

Neural group testing has two key components: (1) \textbf{Neural Group Testing Network Design}, which modifies a neural network so that it could test multiple samples in one forward pass. The group testing network takes a group of samples as input, and learns to predict whether there exists any positive samples in the group. (2) \textbf{Neural Group Testing Algorithm}, which schedules the use of the neural group testing network. Our neural group testing settings are closely related to the sparsity-constrained group testing~\cite{DBLP:journals/tit/GandikotaGJZ19}, where the group size $M$ could not be arbitrarily large. In general, groups of samples that test negative are ruled out. If a group tests positive, samples in that group are then retested adaptively.

\subsection{Neural Group Testing Network Design} 
We consider binary classification tasks with imbalanced class distribution. Suppose that we are given an individual testing network $\varphi: \mathfrak{X} \rightarrow \mathfrak{Y}$ that could only test each sample $x_m \in \mathfrak{X}$ individually, and output a binary label $y_m\in \mathfrak{Y}$ (either positive or negative). Our goal is to modify $\varphi$ into a group testing network $\bPhi$. The input of $\bPhi$ is a set of $M$ samples $X=\{x_1,\ldots,x_M\}, x_m\in \mathfrak{X}$. In general, each group of $M$ samples are sampled uniformly from the entire pool of $N$ data-points. The output of $\bPhi$ should be positive as long as one or more of the set samples are positive.  Specifically, suppose that $y_m$ is the ground-truth label for testing $x_m$ individually, then the ground-truth label for testing $X$ is $ \yy = \max_{1 \leq m \leq M} y_m$. In addition, we assume a training set is available for fine-tuning the group testing network $\bPhi$.

\subsubsection{Design 1: Merging Samples in Pixels} 
We first present a simple design that does not require any modification to the individual testing network, \ie $\bPhi=\varphi$, inspired by \textit{Mixup} data augmentation method for image classification tasks~\cite{mixup}. Given an original training set, \textit{Mixup} generates a new training set by averaging $M$ random images in the pixel space, as well as their labels. For example, if $M=2$ and the two random images $x_1$, $x_2$ are a cat image and a dog image respectively, then the new image will be $(x_1+x_2)/2$ and the corresponding label will be ``$0.5$ cat and $0.5$ dog''. \textit{Mixup} suggests that training on the new training set with mixed samples and mixed labels serves the purpose of data augmentation, and achieves better test accuracy on normal images. Inspired by \textit{Mixup}, Design 1 tests a group of $M$ samples $X=\{x_1,\ldots,x_M\}, x_m\in \mathfrak{X}$ by simply averaging them in the pixel space into a mixed sample $\frac{1}{M} \left( \sum_{x\in X} x_m \right)$ (See Figure~\ref{fig:Design}, Design 1). We fine-tune $\bPhi$ using the train set so that $\bPhi$ learns to predict positive if there is any positive sample in a group. We note that though simple, the design already saves a considerable amount of computation: Even with the smallest group size $M=2$, Design 1 could reduce the computation cost by nearly half ideally.

\subsubsection{Design 2: Merging Samples in the Feature Space}

Design 2 considers a general strategy for pooling samples with a neural network. We consider the general architecture of group testing network $\bPhi$ that directly takes a set of $M$ samples as the input. We want $\bPhi$ to be a well-defined function on sets. This requires $\bPhi$ to be permutation invariant to the ordering of the set samples, \ie for any permutation $\pi$: 
\begin{equation*}
\bPhi(\{x_1,\ldots,x_M\}) = \bPhi(\{x_{\pi(1)},\ldots,x_{\pi(M)}\})
\end{equation*}
Regarding the permutation invariant property of a neural network, Zaheer \etal~\cite{deepsets} proves the following theorem:

\begin{theorem}
~\cite{deepsets}
Assume the elements are countable, 
i.e. $|\mathfrak{X}| < \aleph_0$. 
A function $h:2^\mathfrak{X}\to\mathbb{R}$ operating on a set $X$ is \textbf{invariant} to the permutation of elements in $X$ if and only if it can be decomposed in the form $h(X) = \rho \left( \sum_{x\in X} \phi(x) \right)$, for suitable transformations $\phi$ and $\rho$. 
\end{theorem}

The structure of permutation invariant functions in Theorem 1 hints a general strategy for designing a neural network that is a well-defined function on sets. The neural network has the following three steps: 

\begin{itemize} [leftmargin=3mm, itemsep=0mm,partopsep=0pt,parsep=0pt]

    \item Transform each sample $x_m$ into a feature vector $\phi(x_m)$. 

    \item Aggregate (e.g., sum, average) all feature vectors $\phi(x_m)$. 
    
    \item Use $\rho$ to process the aggregation of the feature vectors.
\end{itemize}

The key of the aforementioned three-step procedure is to insert a feature aggregation operator (Step 2) in the middle of the network. Zaheer \etal~\cite{deepsets} further prove that the aggregation operator could be any permutation-equivariant operator, including max pooling, min pooling. Following the structure of permutation invariant functions in Theorem 1, we now modify the individual testing network $\varphi: $ into a group testing network $\bPhi$ (See Figure~\ref{fig:Design}, Design 2). Without loss of generality, we view $\varphi$ as a composition of a series of nonlinear operations, each represented by a neural layer $f^{(i)}$. 
\begin{equation*}
\varphi (x_m) = f^{(L)} \circ \ldots \circ f^{(2)} \circ f^{(1)} (x_m) 
\end{equation*}
where $\circ$ denotes composition of two functions and $L$ is number of neural layers of $\varphi$. We first use the first $T$ layers in $\varphi$ to map each sample into a feature vector. We then aggregate the feature vectors, and run the following $(L-T)$ layers on the aggregated feature vector. Formally, 
\begin{align*}
\bPhi(X) &= \bPhi(\{x_1,\ldots,x_M\}) = \rho \left( \mathop{aggregate}\limits_{1 \leq m \leq M} \phi(x_m)\right) \\ 
\phi &= f^{(T)} \circ \ldots \circ f^{(2)} \circ f^{(1)} \\
\rho &= f^{(L)} \circ \ldots \circ f^{(T+2)} \circ f^{(T+1)}
\end{align*}

We analyze the computation cost per sample in one forward pass. Suppose that the computation cost of running each layer $f^{(i)}$ is $c_i$. The cost of testing each sample individually is $\sum_{i=1}^{L} c_i$. Design 2 reduces the cost to $\sum_{i=1}^{T} c_i + \frac{1}{M} \sum_{i=T+1}^{L} c_i $ since the computation cost of running the last $(L-T)$ layers are shared by $M$ samples. In addition, we note that Design 1 could be viewed as a special case of Design 2 when $T=0$. In this case, $\phi(x_m)=x_m$ is an identity function, and thus the sample merging happens in the pixel space.

\subsubsection{Design 3: Merging Samples Hierarchically} 

Design 3 merges samples hierarchically and recursively at different levels of the network, leading to a tree-structured merging scheme (See Figure~\ref{fig:Design}, Design 3). Consider a binary tree with $M$ leaf nodes, each corresponding to one sample. We first merge pairs of samples, aggregating $M$ leaf nodes into $\frac{M}{2}$ level-$1$ branch nodes. We run several neural layers to transform each branch node's feature, and then further aggregates $\frac{M}{2}$ level-$1$ branch nodes into $\frac{M}{4}$ level-$2$ branch nodes. We repeat the process until we get the root node, and we run the rest of neural layers on the root node's feature to get the prediction. Formally, assumes that the group size $M$ is a power of two. We use $h_{i,j}$ to denote the feature of the $j^{th}$ node at level-$i$, where $i = 0, \dots, \log_2{M}$. 
\begin{align*}
h_{0,j} &= x_j, \quad j = 1, 2, \dots, M  \\ 
h_{i+1,j} &= \mathop{aggregate}( \phi^{(i)}(h_{i,2j-1}) , \phi^{(i)}(h_{i,2j}))\\ 
\bPhi(X) &= \bPhi(\{x_1,\ldots,x_M\}) = \rho \left( h_{0,\log_2{M}} \right) \\ 
\phi^{(0)} &= f^{(T_{0})} \circ \ldots \circ f^{(2)} \circ f^{(1)} \\
\phi^{(i+1)} &= f^{(T_{i+1})} \circ \ldots \circ f^{(T_{i}+2)} \circ f^{(T_{i}+1)}\\
\rho &= f^{(L)} \circ \ldots \circ f^{(T_{\log_2{M}}+2)} \circ f^{(T_{\log_2{M}}+1)} 
\end{align*}
where $h_{0,\log_2{M}}$ denotes the root node and $\phi^{(i)}$ denotes the neural layers for transforming each branch node's feature at level-$i$. We build $\phi^{(i)}$ by re-using the neural layers in the individual testing network $\varphi$. We note that Design 3 is not limited to binary trees, and applications can set different fan-out values for different branch nodes in the tree. Finally, we note that Design 2 could be viewed as a special case of Design 3 where the tree is an $M$-ary tree of height $1$. In this case, there is only one level of feature aggregation.

\subsubsection{Implementation and Training Procedure}
For all aforementioned designs, we initialize the parameters of the group testing network $\bPhi$ using the corresponding parameters in the individual testing network $\varphi$. We fine-tune~\cite{DBLP:conf/acl/LiangZY20,DBLP:conf/emnlp/LiangZY20} the group testing network $\bPhi$ on an annotated training set as follows: In each training epoch, we randomly sample an equal number of positive groups (\ie containing at least one positive sample) and negative groups of $M$ samples from the entire pool of $N$ data-points as the training data. We then fine-tune the group testing network $\bPhi$ using the sampled groups, as well as the label corresponding to each group (\ie whether the group contains at least one positive sample). 

\subsection{Neural Group Testing Algorithm}
In neural group testing, standard group testing algorithms can be plugged in to schedule the use of the group testing network. We briefly introduce three standard group testing algorithms. We refer interested readers to \cite{du2000combinatorial,DBLP:conf/colt/Coja-OghlanGHL20} for the theoretical analysis and more sophisticated versions.

\subsubsection{Algorithm 1: Two-Round Testing}
The two-round testing algorithm was first proposed in the Second World War to test soldiers for syphilis~\cite{dorfman1943detection}. As shown in Figure~\ref{fig:algorithmSmall}, this algorithm first randomly splits the data in equal-sized groups, and perform tests group by group. Groups of samples that test negative are ruled out. If a group tests positive, every sample is retested individually. Similarly, in our neural group testing setting, we conduct two rounds of testing: The first round tests samples in groups of size $M$ using the group testing network $\bPhi$. If a group tests positive, we use the individual testing network $\varphi$ to test each sample in positive groups individually.

\begin{table*}[ht]
\small
\resizebox{\linewidth}{!}{
\tabcolsep=0.11cm
\begin{tabular}{ccz{2}rrrrccc}
\cmidrule[\heavyrulewidth]{1-10}
\multicolumn{3}{c}{ \bf Configuration } & \multicolumn{2}{c}{ \bf Performance } & \multicolumn{5}{c}{ \bf Computational Efficiency } \\
\cmidrule(lr){1-3}
\cmidrule(lr){4-5}
\cmidrule(lr){6-10}
\bf Algorithm & \bf Design & 
\multicolumn{1}{c}{\textbf{\begin{tabular}[c]{@{}c@{}}Group \\ Size $M$\end{tabular}}}
& 
\bf Recall 
& 
\multicolumn{1}{c}{\textbf{\begin{tabular}[c]{@{}c@{}}False \\ Positive \\ Rate\end{tabular}}}
& 
\multicolumn{1}{c}{\textbf{\begin{tabular}[c]{@{}c@{}}\# of Test\\ 1st round \end{tabular}}}
& 
\multicolumn{1}{c}{\textbf{\begin{tabular}[c]{@{}c@{}} \# of Test\\ Total \end{tabular}}}
& 
\multicolumn{1}{c}{\textbf{\begin{tabular}[c]{@{}c@{}} 
\# of Individual 
\\ Results Obtained \\ per Test Used  \end{tabular}}}
& 
\multicolumn{1}{c}{\textbf{\begin{tabular}[c]{@{}c@{}} Total \\ Computation \\ (TMAC) \end{tabular}}}
& 
\multicolumn{1}{c}{
\textbf{\begin{tabular}[c]{@{}c@{}}Computation \\ Relative to \\ Individual Testing
\end{tabular}}
}
\\
\cmidrule{1-10}
\multicolumn{2}{c}{ Individual Testing } & 1 & 100\% & 0.18\% & 48,800 & 48,800 & 1.00 & 805.2 & 100.00\% \\
\cmidrule{1-10}
\multirow{9}{*}{ 
\begin{tabular}[c]{@{}c@{}}Algorithm 1\\ (Two-Round) \end{tabular}
} & \multirow{2}{*}{ 
\begin{tabular}[c]{@{}c@{}}Design 1\\ (Pixel Merge) \end{tabular}
} & 2 & \cellcolor{red!10} 92\% & \cellcolor{green!10} 0.08\% & 24,400 & 27,394 & \cellcolor{green!10} 1.78 & 452.0 & \cellcolor{green!10} 56.14\% \\
& & 4 & \cellcolor{red!40} 64\% & \cellcolor{green!10} 0.07\% & 12,200 & 18,864 & \cellcolor{green!15} 2.59 & 311.3 & \cellcolor{green!30} 38.66\% \\
\cmidrule{2-10}
& 
\multirow{4}{*}{ 
\begin{tabular}[c]{@{}c@{}}Design 2\\ (Feature Merge) \end{tabular}
} & 2 & 100\% & \cellcolor{green!30} 0.02\% & 24,400 & 24,634 & \cellcolor{green!10} 1.98 & 495.8 & \cellcolor{green!10} 61.57\% \\
& & 4 & 100\% & \cellcolor{green!30} 0.03\% & 12,200 & 12,980 & \cellcolor{green!25} 3.76 & 347.9 & \cellcolor{green!20} 43.20\% \\
& & 8 & 100\% & \cellcolor{green!20} 0.04\% & 6,100 & 8,356 & \cellcolor{green!35} 5.84 & 293.8 & \cellcolor{green!30} 36.49\% \\
& & 16 & 100\% & \cellcolor{green!10} 0.07\% & 3,050 & 9,338 & \cellcolor{red!10} 5.23 & 321.1 & \cellcolor{green!30} 39.88\% \\
\cmidrule{2-10}
& \multirow{3}{*}{ 
\begin{tabular}[c]{@{}c@{}}Design 3\\ (Tree Merge) \end{tabular}
} & 4 & 100\% & \cellcolor{green!30} 0.03\% & 12,200 & 12,908 & \cellcolor{green!25} 3.78 & 292.9 & \cellcolor{green!30} 36.38\% \\
& & 8 & 100\% & \cellcolor{green!20} 0.05\% & 6,100 & 9,308 & \cellcolor{green!35} 5.24 & 255.8 & \cellcolor{green!30} 31.76\% \\
& & 16 & 100\% & \cellcolor{green!10} 0.08\% & 3,050 & 15,626 & \cellcolor{red!20} 3.12 & 371.1 & \cellcolor{green!20} 46.09\% \\
\cmidrule{1-10}
\multirow{4}{*}{ 
\bf 
\begin{tabular}[c]{@{}c@{}}Algorithm 2\\ (Multi-Round) \end{tabular}
} & \multirow{2}{*}{
\begin{tabular}[c]{@{}c@{}}Design 2\\ (Feature Merge) \end{tabular}
} & 8 & 100\% & \cellcolor{green!30} 0.01\% & 6,100 & 7,392 & \cellcolor{green!30} 6.60 & 273.8 & \cellcolor{green!30} 34.00\% \\
& & 16 & 100\% & \cellcolor{green!30} 0.02\% & 3,050 & 5,130 & \cellcolor{green!50} 9.51 & 246.0 & \cellcolor{green!35} 30.55\% \\
\cmidrule{2-10}
& \multirow{2}{*}{
\bf 
\begin{tabular}[c]{@{}c@{}} Design 3\\ (Tree Merge) \end{tabular}
} & \bf 8 & \bf 100\% & \bf \cellcolor{green!30} 0.01\% & \bf 6,100 & \bf 7,844 & \bf \cellcolor{green!30} 6.22 & \bf 225.8 & \bf \cellcolor{green!50} 28.04\% \\
& & \bf 16 & \bf 100\% & \bf \cellcolor{green!30} 0.02\% & \bf 3,050 & \bf 6,726 & \bf \cellcolor{green!45} 7.26 & \bf 212.8 & \bf \cellcolor{green!50} 26.43\% \\
\cmidrule{1-10}
\begin{tabular}[c]{@{}c@{}}Algorithm 3\\ (One-Round) \end{tabular}
& 
\begin{tabular}[c]{@{}c@{}}Design 3\\ (Tree Merge) \end{tabular}
& 4 & 100\% & \cellcolor{green!10} 0.11\% & 24,400 & 24,400 & \cellcolor{green!10} 2.00 & 491.8 & \cellcolor{green!10} 61.08\% \\
\cmidrule[\heavyrulewidth]{1-10}
\end{tabular}
} 
\caption{
Computational savings and performance improvement achieved by neural group testing with various configurations. ``\# of Individual Results Obtained per Test Used'' is the total number of forward passes divided by the number of samples in the test set. Evaluated on the firearm image moderation task with prevalence $0.1\%$. 
}
\label{tab:mainresult}
\end{table*}

\subsubsection{Algorithm 2: Multi-Round Testing}
Algorithm 2 improves Algorithm 1 by adding further rounds of group tests, before testing each sample separately~\cite{du2000combinatorial} (Figure~\ref{fig:algorithmSmall}). If a group of $M$ samples tests positive, we split the group into $K$ sub-groups of size $M/K$ and tests the sub-groups recursively. Sub-groups of samples that test negative are ruled out. One way to implement Algorithm 2 in neural group testing is to train multiple versions of the group testing network $\bPhi$, each corresponding to a possible group size ($M, M/K,...$). In practice, we find that the group testing network $\bPhi$ trained with group size $M$ also works well for smaller group sizes (\eg $M/K$). Therefore, we use the same group testing network $\bPhi$ for all rounds except the individual testing round. 
Using the same group testing network for different rounds provides an additional computational benefit for Design 2 \& Design 3: We can cache samples' features  $\phi(x_m)$ to achieve more computation saving.

\subsubsection{Algorithm 3: One-Round Testing}
Algorithm 1 \& Algorithm 2 saves computation, but increases the tail latency since there are multiple rounds of tests to identify a positive sample. Algorithm 3 performs all the tests in one round, with many overlapping groups~\cite{doublePoolTest,DBLP:journals/corr/abs-2005-06617}, and then decodes the test results. Algorithm 3 is also known as non-adaptive group testing. In our neural group testing, we use a simple version of non-adaptive group testing: double pool testing~\cite{doublePoolTest}. We first split the data into groups of size $M$ twice, so that each sample occurs in exactly two groups. We then decode the results by ruling-out the items that occur at least once in a negative group. 

\section{Experiments}
\label{sec:experiment}

\subsection{Application and Dataset} 
We consider image moderation~\cite{moderator}, a task to identify and remove rare but inappropriate images from a large number of user-uploaded images. AI-based image moderation has been widely deployed in Facebook, YouTube to scale with the large number of uploaded images~\cite{cambridgeconsultants}. For example, Facebook receives $350$ million uploaded photos every day~\cite{facebookUpload}. The global image moderation solution market reached a value of USD $4.9$ billion in 2019 and is expected to reach a valuation of nearly USD $8.9$ billion by 2025~\cite{contentModerationMarket}. In this paper, We focus on detecting images of firearms. Gun violence is always a hot topic. Legit or not, pictures showing someone posing with a firearm can offend visitors, shock young minds, and displease advertisers~\cite{PicPurify}.

We build a firearm moderation dataset based on ImageNet~\cite{imagenet}. We construct a large image test set with around 50k images and 0.1\% firearm images using the validation set in ImageNet, Following the reported prevalence of violent images in Facebook~\cite{facebookModeration}, We exclude ambiguous classes like ``military uniform'', ``holster'' from ImageNet which might contain firearms. We use the train set in ImageNet for training the neural group testing network. The code for generating the dataset is available in our Github repo. Evaluating neural group testing on other unstructured data types like text is an interesting direction of future work~\cite{DBLP:conf/aaai/LiangTCY20,liang2021herald}.

\subsection{Setup}
We implement the individual testing network $\varphi$ with ImageNet pre-trained ResNext-101~\cite{resnext} with \textit{Mixup} data augmentation. 
We use maxpool to aggregate samples, and set $T=0.2L$ for Design 2\&3. 
We experiment with various group sizes $M=2,4,8,16$, as well as different design choices of neural group testing network and neural group testing algorithm. 
We report the number of multiply-accumulate operations (MAC) during inference measured by \textit{ptflops}~\cite{ptflops}, as well as the detection performance in terms of recall and false positive rate.

\subsection{Results}

\subsubsection{Performance Analysis} Interestingly, group testing improves detection performance. Table~\ref{tab:mainresult} shows that all configurations have a lower false positive rate than individual testing. This is because neural group testing performs multiple tests before confirming that a sample is positive. In addition, all configurations except Design 1 achieves 100\% recall, showing that merging samples' pixels is not ideal and it is necessary to merge samples in the feature space.

\subsubsection{Computational Efficiency Analysis} 
As shown in Table~\ref{tab:mainresult}, our best configuration (Algorithm 2 + Design 3 + Group Size 16) reduces the overall computation cost by over $73\%$ (relative computation $26.43\%$). If we switch from Algorithm 2 to Algorithm 1, the relative computation increases a lot ($26.43\%$ vs. $46.09\%$) and the number of results obtained per test ($3.12$) is far smaller than the group size $16$. The reason is that, though a large group size saves computation in the first round, it passes many samples to the following rounds since each positive group contains more samples. Therefore, it is important to use multiple rounds of testing (Algorithm 2). In addition, switching from Design 3 to Design 2, the relative computation also increases significantly ($26.43\%$ vs. $30.55\%$), showing that Design 3 also contributes to the computation savings.

\subsubsection{Various Prevalence Levels} 
\begin{figure}[tb]
\centering
\includegraphics[width=0.473\textwidth]{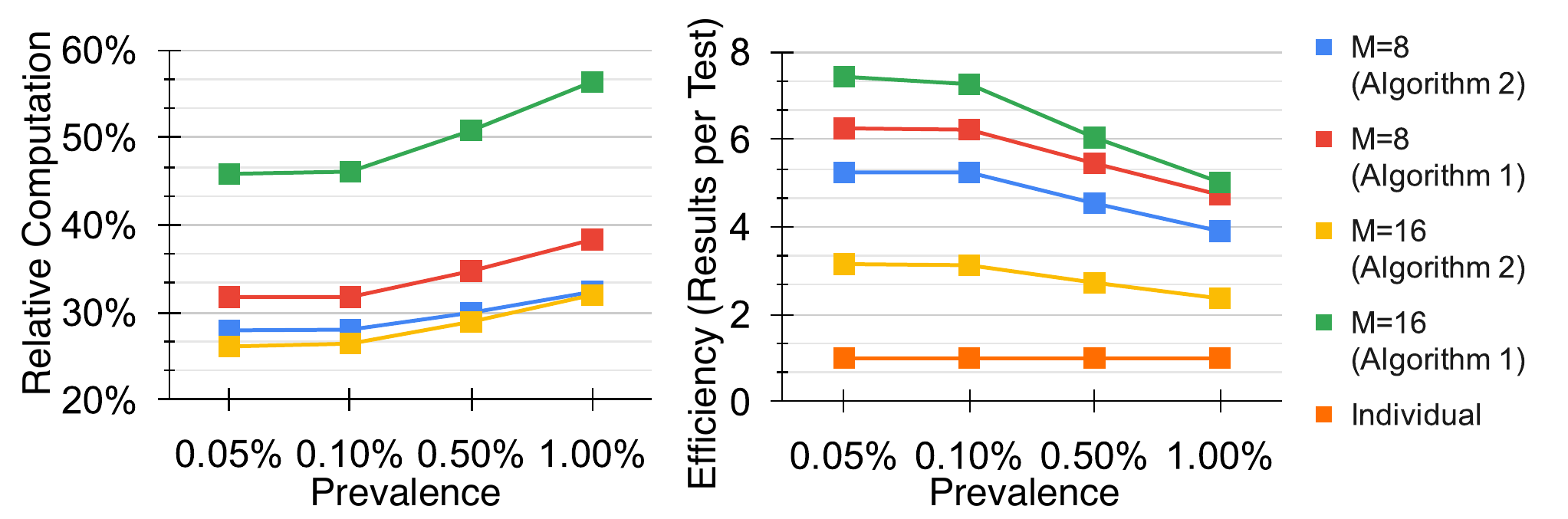} 
    \caption{
    Relative computation and test efficiency 
    under different prevalence.
    Evaluated with Design 3 (Tree Merge) with various group size $M$ and algorithms. 
    } 
    \label{fig:chart}
\end{figure}

Since Design 3 achieves the best computational efficiency, we experiment with different levels of prevalence with Design 3. We change the prevalence by randomly sampling and transforming the firearm images. Figure~\ref{fig:chart} shows that, our method is able to reduce the relative computation to less than $33\%$ across all settings. As expected, neural group testing is most efficient when the prevalence is low, because group tests are more likely to be negative, which saves testing many samples individually.

\subsubsection{Combining with Existing Acceleration Methods} 

Because neural group testing can be performed without modifying the neural network architecture, it is complementary to and can be combined with all the popular approaches to speed up inference, such as sparsifying or quantizing networks or downsampling images. To show this, we apply neural network pruning on our best configurations (Algorithm 2 + Design 3 + Group Size $M=8,16$). Following~\cite{pruning}, we prune the group testing network $\bPhi$ by removing the weights with the smallest absolute values. We first sort the parameters in the neural network with their bsolute values, and then sparsify $80\%$ of the parameters with low absolute values. We evaluated the pruned version of the group testing network $\bPhi$ in our neural group testing settings and found no detection performance drop. This shows that neural group testing can be combined with other existing approaches to speed up inference. 
 
\section{Conclusion}

In this paper, we present neural group testing, a general deep learning acceleration framework that tests multiple samples in one forward pass. We found that multi-round tree merge is the best design for neural group testing. It can group up to $16$ images in one forward pass and reduce the overall computation cost by over $73\%$ while improving detection performance. Another benefit of this design is that it can be easily combined with other approaches to accelerate inference by, for example, quantizing or pruning network parameters or downsamping input data. We applied existing neural network pruning methods, and found that we can sparsify $80\%$ of the parameters in the neural network without any performance drop in neural group testing. Evaluating further gains by combining other orthogonal approaches is an interesting direction of future work.

\section*{Acknowledgments}
We thank ISIT 2021 chairs and reviewers for their helpful feedback. We also thank Mingtao Xia for helpful discussions on group testing. This research is supported by NSF CAREER 1942926 and grants from the Chan-Zuckerberg Initiative. 


\bibliographystyle{IEEEtran} 
\bibliography{ref}


\end{document}